\documentclass[conference]{IEEEtran}
\usepackage{cite}
\usepackage{amsmath,amssymb,amsfonts,fourier}
\usepackage{algorithmic}
\usepackage{graphicx}
\usepackage{textcomp}
\usepackage{xcolor}
\usepackage{subcaption}
\usepackage{enumitem}
\def\BibTeX{{\rm B\kern-.05em{\sc i\kern-.025em b}\kern-.08em
    T\kern-.1667em\lower.7ex\hbox{E}\kern-.125emX}}
\usepackage{hyperref}

\newcommand{\bx}{\boldsymbol{x}}
\newcommand{\by}{\boldsymbol{y}}
\newcommand{\bh}{\boldsymbol{h}}
\newcommand{\bbeta}{\boldsymbol{{\beta}}}
\newcommand{\bphi}{\boldsymbol{{\phi}}}
\newcommand{\bpsi}{\boldsymbol{{\psi}}}
\newcommand{\ndim}{{d}}
\newcommand{\Rf}{\mathbb{R}}
\newcommand{\N}{\mathbb{N}}
\newcommand{\ST}{{\mathcal{S}}}

\newcommand{\defined}{\, := \,}

\renewcommand{\(}{\left(}
\renewcommand{\)}{\right)}
\newcommand{\bdm}{\begin{displaymath}}
\newcommand{\edm}{\end{displaymath}}
\newcommand{\transp}{{\scriptscriptstyle{\mathsf{T}}}}

\begin{document}

\title{Explainable and Class-Revealing Signal Feature Extraction
  via Scattering Transform and Constrained Zeroth-Order Optimization}

\author{\IEEEauthorblockN{Naoki Saito}
\IEEEauthorblockA{\textit{Department of Mathematics} \\
\textit{University of California}\\
Davis, CA 95616 USA \\
nsaito@ucdavis.edu}
\and
\IEEEauthorblockN{David Weber}
\IEEEauthorblockA{\textit{The Delphi Group} \\
\textit{Carnegie Mellon University}\\
Pittsburgh, PA 15213 USA \\
dsweber2@protonmail.com}
}

\maketitle

\begin{abstract}
  We propose a new method to extract discriminant and explainable features from a particular machine learning model, i.e., a combination of the scattering transform and the multiclass logistic regression. Although this model is well-known for its ability to learn various signal classes with high classification rate, it remains elusive to understand why it can generate such successful classification,
  mainly due to the nonlinearity of the scattering transform. In order to uncover the meaning of the scattering transform coefficients selected by the multiclass logistic regression (with the Lasso penalty), we adopt zeroth-order optimization algorithms to search an input pattern that maximizes the class probability of a class of interest given the learned model. In order to do so, it turns out that imposing sparsity and smoothness of input patterns is important. 
We demonstrate the effectiveness of our proposed method using a couple of synthetic time-series classification problems.
\end{abstract}

\begin{IEEEkeywords}
Nonlinear discriminant feature extraction, scattering transform, wavelets, zeroth-order optimization, sparsity and smoothness of signals
\end{IEEEkeywords}

\section{Introduction}
\label{sec:intro}
In signal and image classifications, the \emph{Scattering Transform} (ST)
\cite{MALLAT-SCAT, BRUNA-MALLAT}, which cascades wavelet transform convolutions
with modulus nonlinearities (i.e., absolute values) and averaging operators,
has emerged as an alternative to the popular Convolutional Neural Networks
(CNNs)/Deep Learning (DL)~\cite{LECUN-BENGIO-HINTON, Goodfellow-et-al-2016}.
Since only two or three layers of the cascades in the ST are sufficient and
since it uses the predefined wavelet convolution filters,
it has a number of advantages over CNNs/DL for signal classification problems:
1) its training process is computationally faster;
2) it does not require a large number of training samples;
3) it automatically generates multiscale/multifrequency feature representations
of input data in an \emph{explicit} manner via wavelet filters;
4) the computed ST coefficients can be fed to any classifier of choice, 
e.g., Multiclass Logistic Regression (MLR), Support Vector Machine (SVM), $k$-Nearest Neighbors ($k$NN), etc.; and 
5) it is more mathematically tenable.
Yet, it still extracts robust and quasi-invariant features of input signals
relative to certain types of deformations or perturbations (e.g., a small amount
of shifts, warps, noise, etc.) so that it provides comparable classification
performance as DL models.
In fact, for a small number of training samples, it outperforms DL
models~\cite{BRUNA-MALLAT-CVPR11, BRUNA-MALLAT, ANDEN-MALLAT}.

Our earlier work has obtained excellent results using the ST and its
variant in a number of signal classification problems including:
underwater object classification from acoustic wavefields~\cite{SAITO-WEBER-SPIE, WEBER-PHD}; texture image classification~\cite{CHAK-SAITO-MWSN}; and music
genre classification~\cite{CHAK-SAITO-WEBER}.
However, one critical thing remains to be understood: \emph{Why does this combination of the features and the classifier work so well? What are the features
\underline{in the original time domain} that contributed to these excellent
classification results?} Because the ST is a nonlinear transform like DL, even
if we can identify which ST coefficients significantly contributed to the
correct classification, the true meaning of such coefficients in the original
time domain has been elusive: the only explicitly available information is the
so-called \emph{path} information, i.e., a set of indices of the wavelet filters
(e.g., passband info) used in all the previous layers and the current layer
in order to generate that particular ST coefficient.

Hence, our aim here is to \emph{extract explainable or class-revealing features
using the logistic regression coefficients learned on the ST coefficients of
given training samples.}
More specifically, we plan to \emph{estimate signals that maximize the class
probability using their ST coefficients and the trained MLR classifier for a signal class of interest}.
If we are successful, then we can deepen our understanding of the features
in the original time domain that are responsible for discriminating one class
from the other.
Such understanding and insight may uncover the secret of physical nature of
a given problem, e.g., detection and classification of underwater objects via
scattered acoustic wavefields; see \cite{SAITO-WEBER-SPIE, WEBER-PHD} for
the background information and our previous effort.
This is quite important since it may provide us with an opportunity to
\emph{design new transmitter signal patterns and/or sensors that specifically
focus on discriminating a particular class of objects}.

\section{Our Proposed Method}
\label{sec:method}
To be concrete, let us consider a typical signal classification problem.
Let $\{\bx_i\}_{i=1:N}$ be available training signals each of which has $\ndim$
time samples. Each $\bx_i \in \Rf^\ndim$ has a class label,
$k \in \{1, \ldots, K\}$, i.e., $\bx_i \in X_k$, where $X_k \subset \Rf^\ndim$ is
a space of class $k$ signals.
Our proposed procedure is the following.
\begin{description}[labelsep=0.5em]
\item[Step 1:] Apply the ST to the training samples;
\item[Step 2:] Train the GLMNet classifier (= the MLR classifier with the Lasso penalty~\cite{HASTIE-TIB-WAINWRIGHT}), which can efficiently
  select a small number of the ST coefficients as key features; let $(\alpha_k, \bbeta_k)$, $k=1:K$ be the resulting intercepts and regression coefficient vectors;
\item[Step 3:] Find an input pattern $\hat{\!\bx} \in \Rf^\ndim$ for class $k$
  that minimizes the following
  criterion:
  \begin{equation}
    \label{eqn:opt}
    \hat{\!\bx} = \arg\min_{\bx \in \Rf^\ndim} \frac{1}{p_k(\bx)} + \mu \| \bx \|_1 + \nu \| \nabla \bx \|_2,
  \end{equation}
  where $p_k(\bx) \defined \exp(\alpha_k+\bbeta_k \cdot \ST[\bx])/\sum_{j=1}^{K} \exp(\alpha_j + \bbeta_j \cdot \ST[\bx])$ is the probability of a signal $\bx$ belonging to class $k$ (according to the trained GLMNet classifier); $\ST[\bx]$ is the ST coefficient vector computed from $\bx$; and  $\mu > 0$, $\nu > 0$ are the Lagrange multiplier parameters to be adjusted.
  Note that the minimization of the first term $1/p_k(\bx)$ is clearly equivalent
  to the maximization of $p_k(\bx)$ while the second and third terms promote
  \emph{sparsity} and \emph{smoothness} of $\bx$, respectively.
\end{description}
We now describe the ST a bit more precisely.
Since we only consider the second-layer ST coefficients, we define it for a given
input 1D signal $\bx \in \Rf^\ndim$ as follows:
\begin{equation}
\label{eqn:ST}
\ST[\bx] \defined  \left\{ \rho\( \rho\(\bx \circledast_{r_1} \bpsi^{(1)}_{\lambda_1}\) \circledast_{r_2} \bpsi^{(2)}_{\lambda_2} \) \circledast_{r_a} \bphi_J \right\}_{\lambda_1 \in \Lambda_1; \lambda_2 \in \Lambda_2},
\end{equation}
where 
$\rho$ is the elementwise modulus operator,
$\rho: \by=(y_1, \ldots, y_\ndim)^\transp \in \Rf^\ndim \mapsto |\by| \defined (|y_1|, \ldots, |y_\ndim|)^\transp \in \Rf_{\geq 0}^\ndim$;
$\circledast_{r}$ indicates a 1D circular convolution (typically done via multiplication in the Fourier domain) \emph{followed by subsampling with rate $r \geq 1$};
$\bpsi^{(l)}_{\lambda_l}$, is the mother wavelet filter with frequency-band index $\lambda_l \in \Lambda_l$ for the $l$th layer, $l=1, 2$; and finally $\bphi_J$ is
the father wavelet (i.e., lowpass) filter at scale $J \in \N$. Note that
we allow different subsampling rates for each layer, i.e., $r_1, r_2$ and
for the final averaging process $r_a$.
For the details, see \cite{MALLAT-SCAT, ANDEN-MALLAT, KYMATIO} as well as
\cite[Chap.~3]{WEBER-PHD}.

Step 3 needs a bit more explanation here.
To solve the minimization problem Eq.~\eqref{eqn:opt}, we use the so-called
\emph{zeroth-order (ZO) or derivative-free optimization}~\cite{CONN-ETAL_ZO-BOOK, LARSON-ETAL_ZO, LIU-ETAL_ZO}. For our proposed method, it is better to use ZO
optimization algorithms than the popular first-order (FO) optimization algorithms
that require computing the gradient of the objective function to be minimized.
This is because the gradient $\nabla \ST[\bx]$ is highly discontinuous, leading
update steps to converge slowly, as shown in \cite[Chap.~4]{WEBER-PHD}.
In this paper, we use the popular ZO method called \emph{Differential Evolution} (DE)~\cite{AHMAD-etal_DE-Rev, BILAL-etal_DE-Rev, OPARA-ARABAS_DE-Rev}.
We also note that we perform the updates of solution candidates in the frequency
domain (via DCT) with the \emph{pink} (a.k.a.\ $1/f$) noise as the initial
randomized candidates instead of time domain updates starting from white noise.
This is because the updates of solution candidates in the time domain
are too local in time and many naturally-occurring signals follow the $1/f$
power spectra~\cite{PRESS}; see also \cite[Chap.~4]{WEBER-PHD}.
The DE is certainly not the only ZO method to use for our problems;
see more discussion of the choice of ZO methods in Section~\ref{sec:summary}.
Finally, we point out the importance of using the sparsity and the smoothness
constraints in Eq.~\eqref{eqn:opt}:
1) the landscape of Eq.~\eqref{eqn:opt} with $\mu=\nu=0$ becomes too rough for the ZO optimization to converge; and 2) the sparsity and the smoothness provide us with easy and intuitive interpretation of signal features.

\section{Examples}
\label{sec:ex}
We now demonstrate the usefulness of our proposed method
using two 
synthetic signal classification problems.

\subsection{``Cylinder-Bell-Funnel'' Signal Classification}
\label{subsec:cbf}
\begin{figure}
  \begin{subfigure}{0.45\textwidth}
  \centering\includegraphics[width=\textwidth]{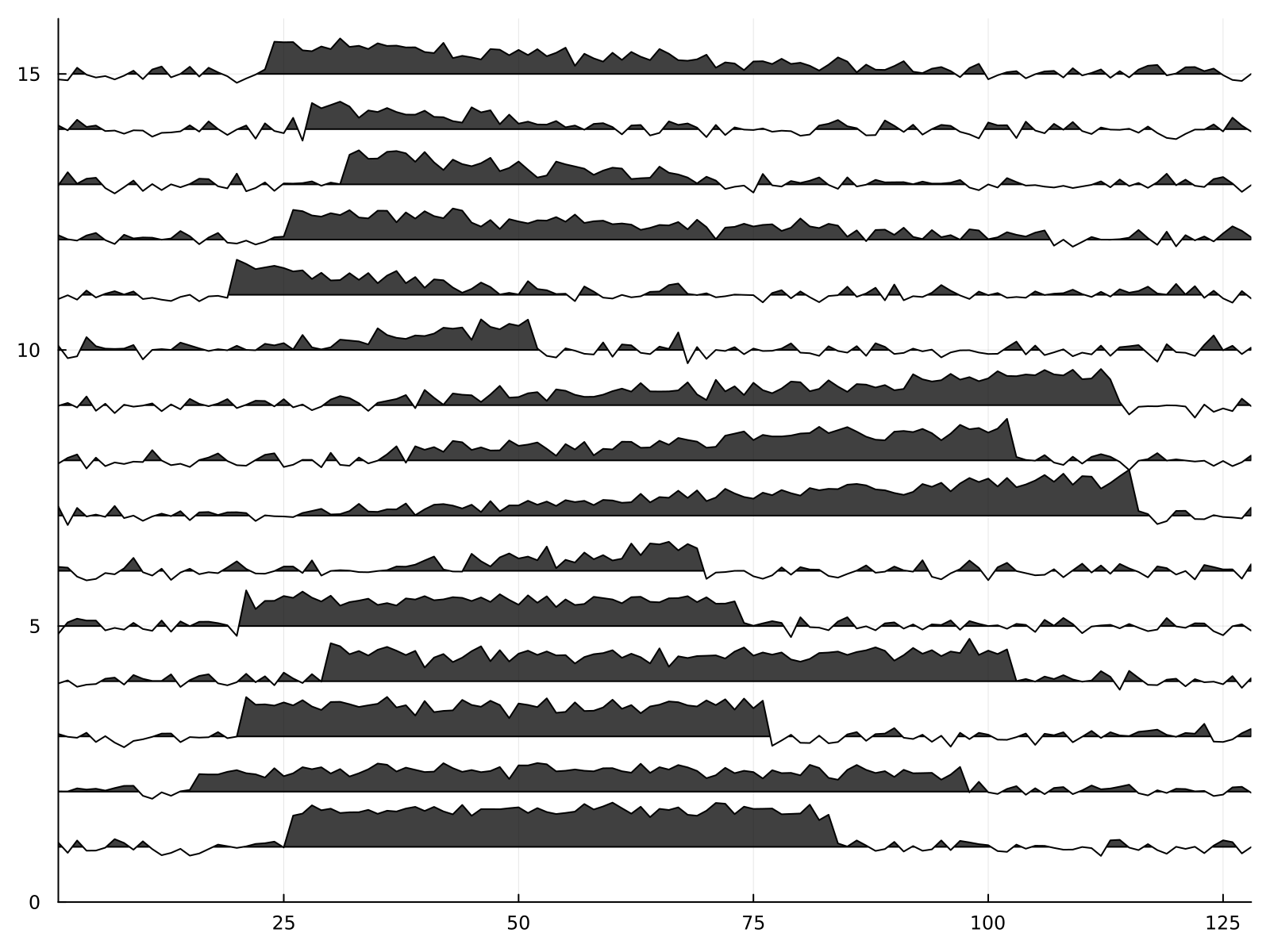}
  \caption{Five samples/class in the ``CBF'' classification problem:
    ``Cylinder'' (bottom 5); ``Bell'' (middle 5); ``Funnel'' (top 5)}
  \label{fig:bcf-samples}
  \end{subfigure}

  \begin{subfigure}{0.45\textwidth}
  \centering\includegraphics[width=\textwidth]{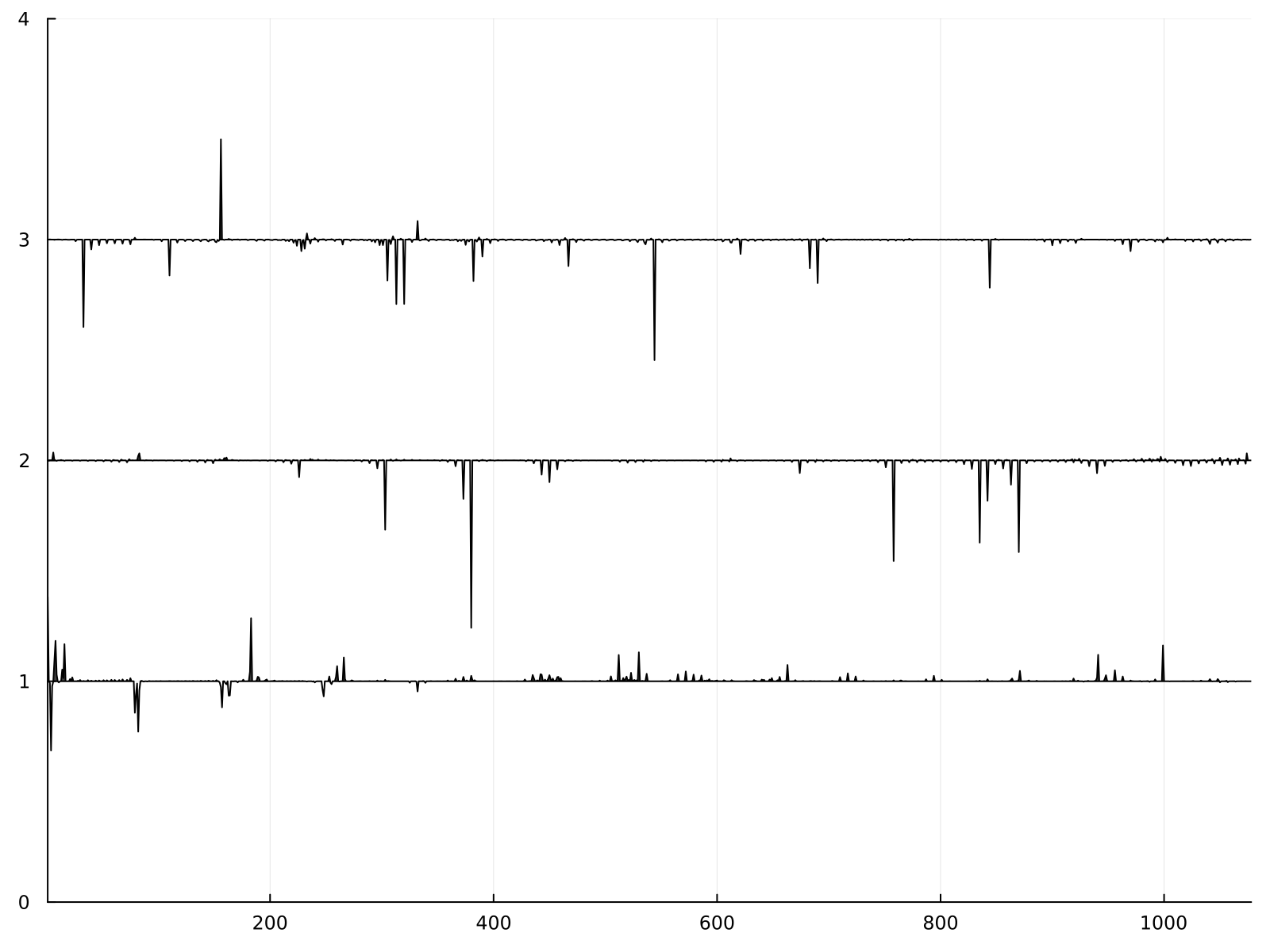}
  \caption{$\bbeta$ coefficients: ``Cylinder'' (bottom);
    ``Bell'' (middle); ``Funnel'' (top)}
  \label{fig:bcf-betas}
  \end{subfigure}

  \begin{subfigure}{0.45\textwidth}
    \centering\includegraphics[width=\textwidth]{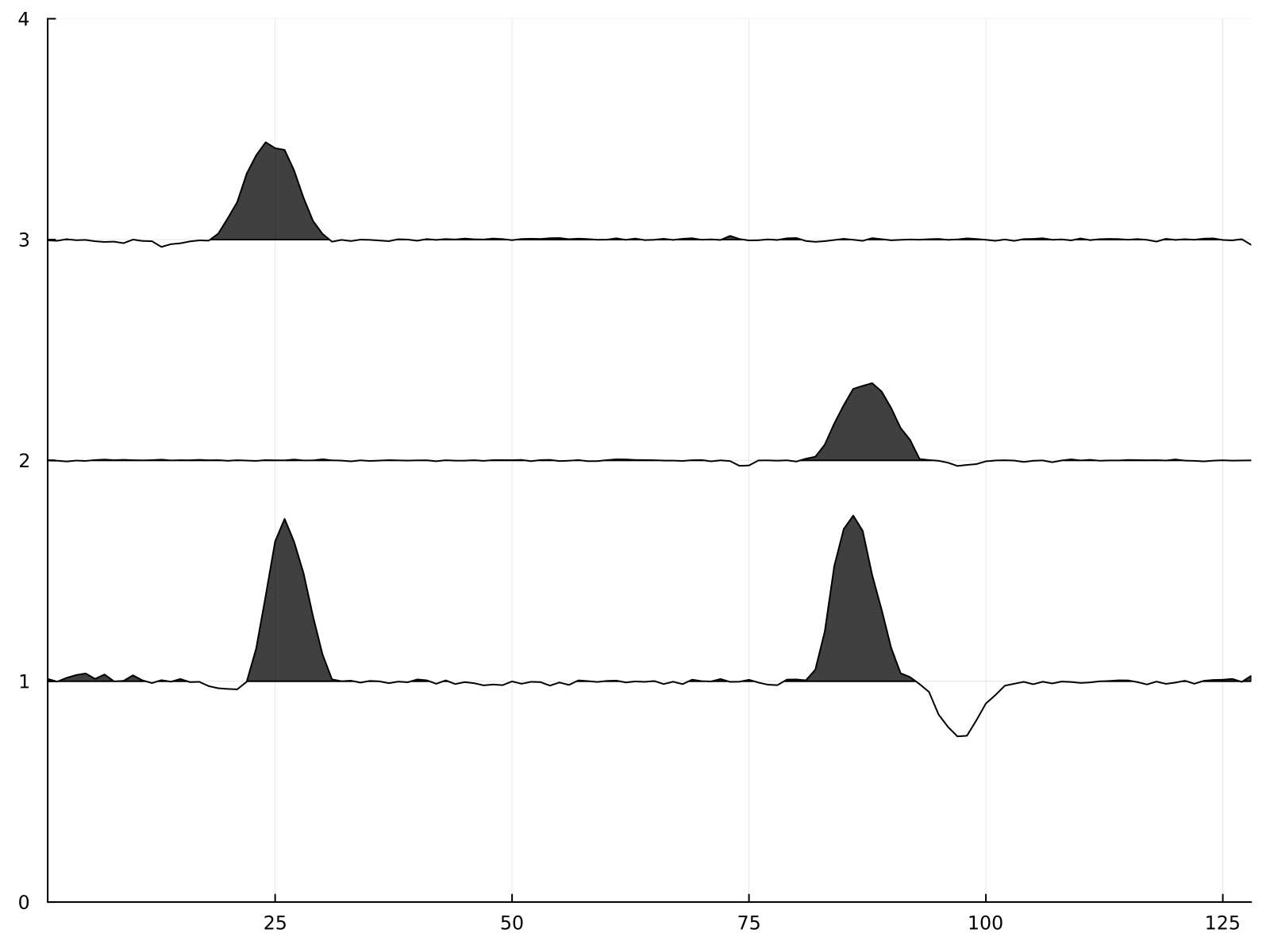}
    \caption{Solutions of Eq.~\eqref{eqn:opt}: ``Cylinder'' (bottom);
    ``Bell'' (middle); ``Funnel'' (top)}

  \label{fig:bcf-sols}
  \end{subfigure}
  \caption{Extracted features via Eq.~\eqref{eqn:opt} reveal the decisive
    characteristics of each class in the ``CBF'' signal classification problem}
\end{figure}
We first tested our idea described above
on a simple yet well-known time-series classification dataset, called
the ``Cylinder-Bell-Funnel'' dataset that was introduced by one of us~\cite{SAITO-COIF-JMIV, SAITO-PHD} and has become quite well known in the time-series
analysis literature; see, e.g., \cite{GEURTS, KEOGH-KASETTY}.
In this example, we want to classify synthetic noisy signals with various shapes,
amplitudes, lengths, and positions into three possible classes ($K=3$).
More precisely, sample signals of the three classes were generated by:
\begin{align*}
c(i) &= (6 + \eta) \cdot \chi_{[a,b]}(i) + \epsilon(i) &&\text{for ``cylinder''}\\
b(i) &= (6 + \eta) \cdot \chi_{[a,b]}(i) \cdot (i-a)/(b-a) + \epsilon(i) && \text{for ``bell''}\\
f(i) &= (6 + \eta) \cdot \chi_{[a,b]}(i) \cdot (b-i)/(b-a) + \epsilon(i) && \text{for ``funnel''}
\end{align*}
where $i=1,\ldots,128$, $a$ is an integer-valued uniform random variable
on the interval $[16,32]$, $b-a$ also obeys an integer-valued uniform
distribution on $[32,96]$, $\eta$ and $\epsilon(i)$ are the standard normal
variates, and $\chi_{[a,b]}(i)$ is the characteristic (or indicator) function
on the interval $[a,b]$.
Figure~\ref{fig:bcf-samples} shows five sample waveforms from each class.
If there is no noise, we can characterize the ``cylinder'' signals by two step
edges and constant values around the center,
the ``bell'' signals by one ramp and one step edge in this order with
positive slopes around the center, and the ``funnel'' signals by
one step edge and one ramp in this order with negative slopes around the center.
In our numerical experiments below, we used the Julia programming language~\cite{JULIA};
more specifically, for the ST computation and the MLR with the Lasso penalty,
we used our own \texttt{ScatteringTransform.jl}~\cite{ScatteringTransform} and
the publicly-available \texttt{GLMNet.jl}~\cite{GLMNet}.
Then, for the DE algorithm, we used the \texttt{BlackBoxOptim.jl}~\cite{BBO} package.

We generated $100$ training signals per class 
and used the ST with the famous ``Mexican-hat'' wavelet function (i.e., the 2nd
derivative of the Gaussian) as its base filter. Finally, we trained the GLMNet
classifier on the 2nd layer ST coefficients $\left\{\ST[\bx_i] \in \Rf_{\geq 0}^{1078}\right\}_{i=1:300}$ of these training signals $\left\{\bx_i \in \Rf^{128}\right\}_{i=1:300}$,
to obtain the regression coefficient vector $\bbeta_k \in \Rf^{1078}$ and
the intercept, $\alpha_k \in \Rf^1$, $k=1, 2, 3$,
corresponding to cylinder, bell, and funnel classes.
The subsampling rates $r_1=r_2=1.5$ and $r_a=8$ were used while
the number of wavelet filters was set as $| \Lambda_1 |=14$ and $|\Lambda_2|=11$.
Hence, the final output size at the 2nd layer of the ST for each input signal
of length $128$ \emph{per filter pair $(\lambda_1, \lambda_2)$} was
$\lfloor 128/1.5/1.5/8 \rfloor = 7$, which led to $7 \times 14 \times 11 = 1078$
ST coefficients in total.
The classification accuracy of this method was almost perfect ($\approx 99$\%)
on newly-generated test dataset of size $3000$.
Figure~\ref{fig:bcf-betas} displays the learned $\bbeta_k$ coefficient vectors
that are quite sparse thanks to the Lasso penalty while Fig.~\ref{fig:bcf-sols}
shows the solutions of Eq.~\eqref{eqn:opt} using the DE algorithm.
It is amazing to see that \emph{those estimated signals pinpoint the
distinguishing features of those three classes}: the one for the bell class and
that for the funnel class are measuring the local activities around the
discontinuous regions where those classes abruptly ends (bell) or starts (funnel)
while it checks \emph{both} edge locations for the cylinder class.
They are the \emph{class-revealing} features and can be viewed as the
\emph{essence} of the prototype signals.

\subsection{Triangular Waveform Classification}
\label{subsec:tri}
\begin{figure}
  \begin{subfigure}{0.45\textwidth}
  \centering\includegraphics[width=\textwidth]{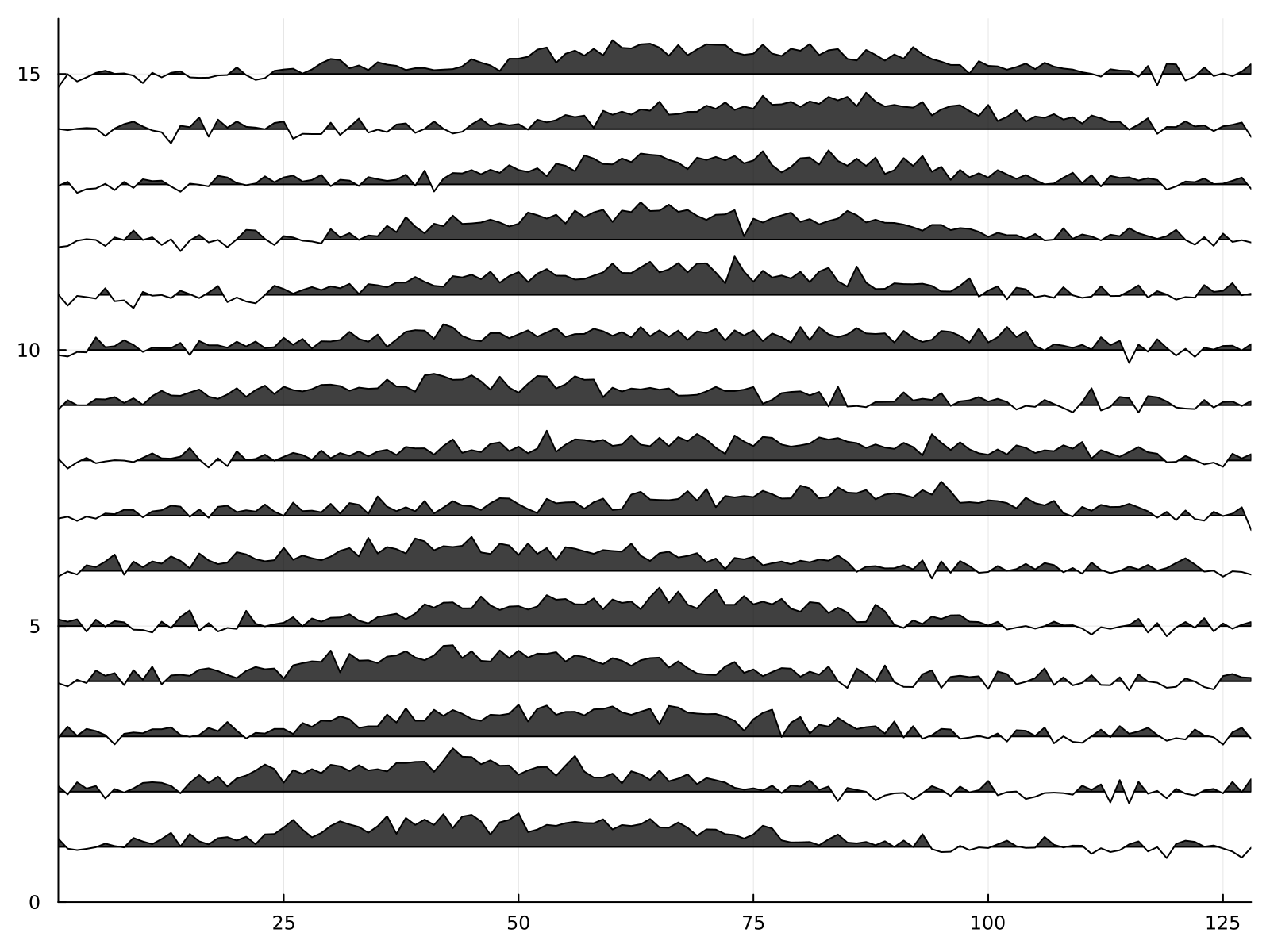}
  \caption{Five samples/class of the triangular waveform problem:
  Class 1 (bottom 5); Class 2 (middle 5); Class 3 (top 5)}
  \label{fig:tri-samples}
  \end{subfigure}

  \begin{subfigure}{0.45\textwidth}
  \centering\includegraphics[width=\textwidth]{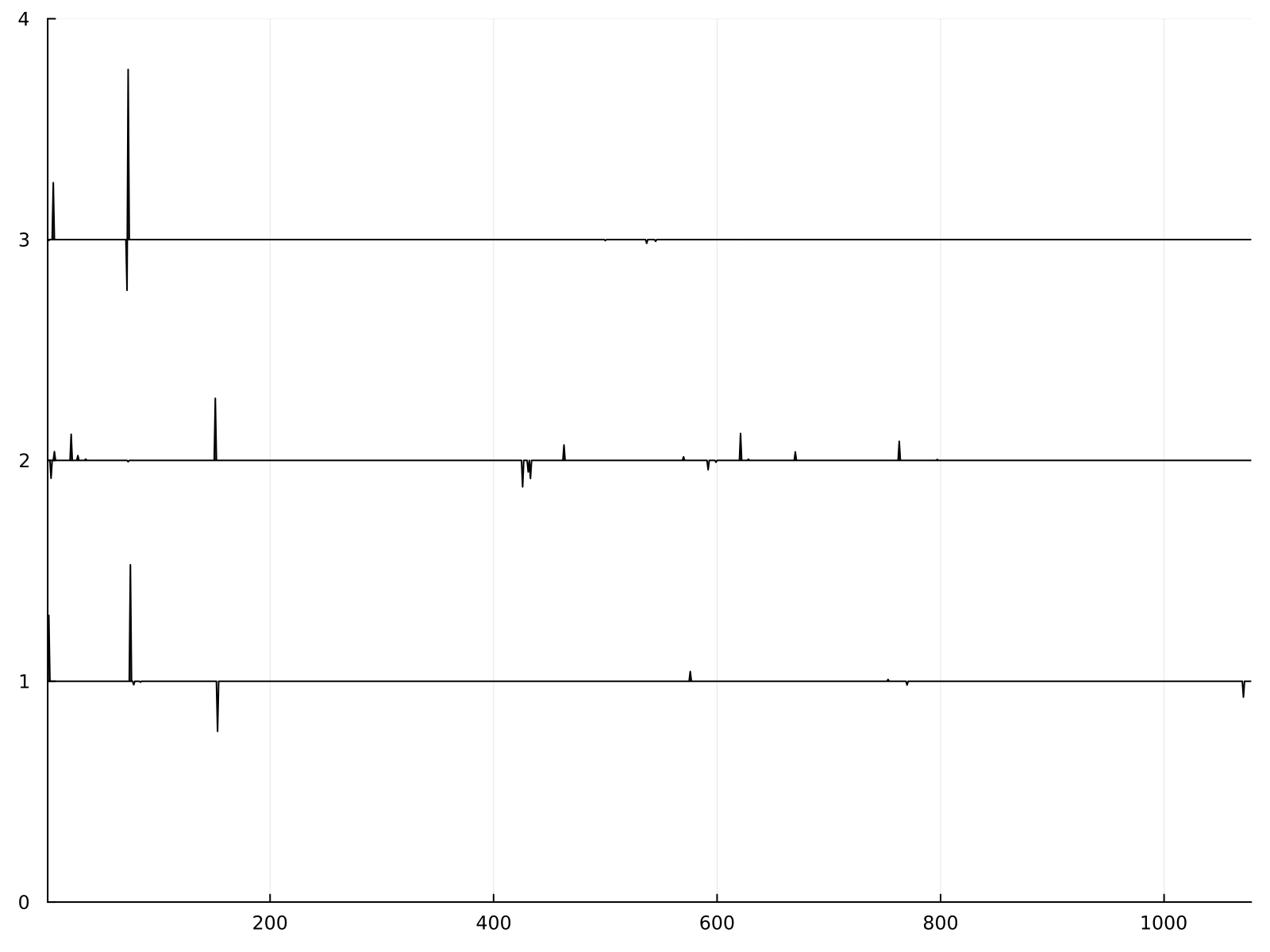}
  \caption{$\bbeta$ coefficients: Class 1 (bottom); Class 2 (middle); Class 3 (top)}
  \label{fig:tri-betas}
  \end{subfigure}

  \begin{subfigure}{0.45\textwidth}
    \centering\includegraphics[width=\textwidth]{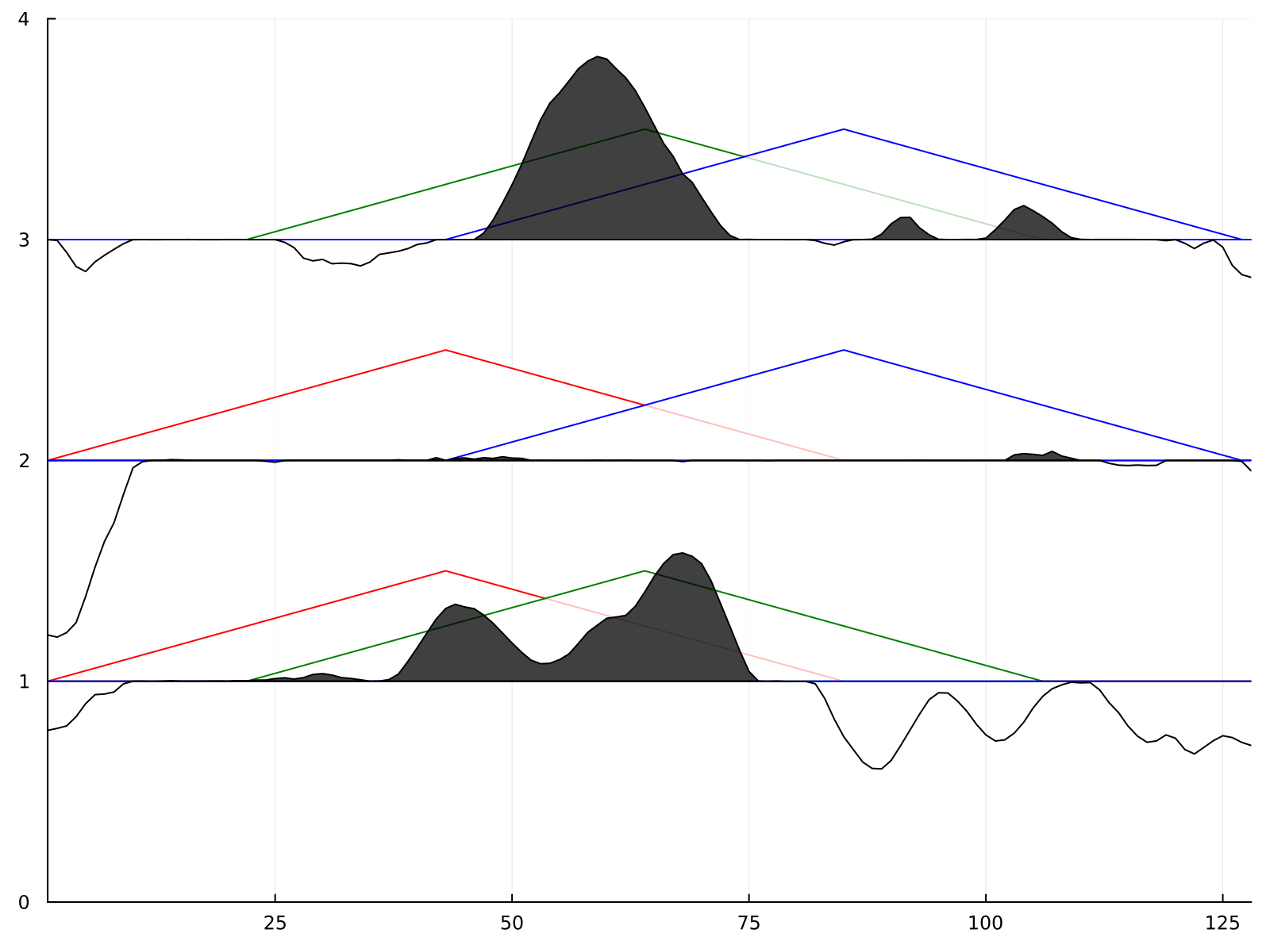}
    \caption{Solutions of Eq.~\eqref{eqn:opt}: Class 1 (bottom); Class 2 (middle); Class 3 (top)}
  \label{fig:tri-sols}
  \end{subfigure}
  \caption{Extracted features via Eq.~\eqref{eqn:opt}
    reveal the significant characteristic of each class
  in the triangular signal classification problem}
\end{figure}
This is another well-known classification problem, first appeared in
the classic book~\cite[Sec.~2.6.2]{CART}.
The dimensionality (or length) of each signal was extended from the original
$21$ in \cite{CART} to $128$ in order to fully utilize the power of the ST:
$21$ was just too short for cascades of wavelet convolutions.
Three classes of signals were generated by: 
\begin{align*}
x^{(1)}(i) &= uh_1(i) + (1 - u)h_2(i) + \epsilon(i) &&\text{for Class 1}, \\
x^{(2)}(i) &= uh_1(i) + (1 - u)h_3(i) + \epsilon(i) &&\text{for Class 2}, \\
x^{(3)}(i) &= uh_2(i) + (1 - u)h_3(i) + \epsilon(i) &&\text{for Class 3},
\end{align*}
where $i = 1, \ldots, 128$, $h_1(i) = \max(6 - |i - 43|/7, 0)$, 
$h_2(i) = h_1(i - 21)$, $h_3(i) = h_1(i - 42)$, $u$ is a uniform random variable
on the interval $(0, 1)$, and $\epsilon(i)$ are the standard normal variates.
Simply speaking, each class consists of random convex linear combination
of two triangles with additive white Gaussian noise.
Notice that the noiseless version of each class forms an edge of
a triangular manifold in $\Rf^{128}$ whose vertices are those three triangles
$\bh_k$, $k=1, 2, 3$. Hence, the noisy versions are distributed within Gaussian
balls around this triangular manifold.
Figure~\ref{fig:tri-samples} shows five sample waveforms from each class;
Fig.~\ref{fig:tri-betas} shows the sparse $\bbeta_k$ coefficient vectors; and
Fig.~\ref{fig:tri-sols} displays the extracted feature/the best input pattern
for each class; the red, green, and blue triangles in the background indicate
the vectors $\bh_1$, $\bh_2$, and $\bh_3$, respectively.
We used the same parameter settings as those of the ``CBF'' signal classification
problem.
The classification rate using the ST plus MLR was $88.9$\% using $1000$ newly generated signals per class as the test dataset while each of the estimated features in Fig.~\ref{fig:tri-sols} gave us essentially perfect classification rate of
$99.99$\%. In other words, these estimated features essentially discovered the
nature of class signal generation mechanism as in the ``CBF'' signal
classification problem.  The Class 1 and the Class 3 features in the bottom and
top rows in Fig.~\ref{fig:tri-sols} indicate the signal energy concentrates
around the apices of $\{\bh_1, \bh_2\}$ and $\{\bh_2, \bh_3\}$, respectively
while the Class 2 feature shown in the middle row of Fig.~\ref{fig:tri-sols}
does not have such energy concentration around $\bh_2$, indicating that the
signals of this class do not have the contribution from $\bh_2$.
This is quite interesting: if the Class 2 feature had energy concentration
around $\{\bh_1, \bh_3\}$, it would have generated worse classification rate
because it would confuse some Class 1 and Class 3 signals; hence our algorithm
decided to use the ``nonexistence'' of the $\bh_2$ component. This has given us
a serendipity: 
in certain situations, it would be better to consider feature ``suppressors''
instead of feature extractors!

\section{Summary and Discussion}
\label{sec:summary}
We have described our effort to understand nonlinear signal features
computed by ST and the MLR classifier 
using the ZO optimization with constraints on sparsity and smoothness
of the features in the original time domain.
Our numerical results 
in Section~\ref{sec:ex}
were quite promising.

We have used the DE algorithm as our ZO method to solve Eq.~\eqref{eqn:opt}.
Clearly, other methods may work well with even faster computational speed than
the DE algorithm. Hence, it is important to evaluate various ZO algorithms that
is robust and computationally efficient.
There are two different strategies in ZO optimization: \emph{single-particle}
methods and \emph{multi-particle} methods.
The advantage of multi-particle methods~\cite{CARRILLO-ETAL_CONSENSUS, FORNASIER-ETAL_CONSENSUS, HUANG-ETAL_PSO} is its population-based adaptive and evolutionary
solution-search capability while the disadvantage is their difficulty to derive
theoretical guarantee of convergence. Hence, we will also investigate the
single-particle methods, which are theoretically more tenable and can fully
utilize the advanced FO 
algorithms once the gradient estimation is done at each iteration. 
In particular, we plan to investigate the so-called \emph{ZO proximal methods}~\cite{POUGKAKIOTIS-KALOGERIAS_ZO-PRIMAL, KAZEMI-WANG_ZO-PRIMAL} because these
can handle the constraints consisting of a nonconvex and smooth function and
a nonsmooth and convex function like our Eq.~\eqref{eqn:opt}.
Our preliminary numerical experiments using the \texttt{PRIMA.jl}~\cite{PRIMA}
is quite promising: it generated similar results as those in Section~\ref{sec:ex}
with an order of magnitude faster than the DE algorithm.

Furthermore, we will investigate:
1) automatic determination of the optimal values of $\mu$ and $\nu$ in
Eq.~\eqref{eqn:opt};
2) other effective constraints beyond sparsity and smoothness;
and
3) the influence of the wavelet filter parameters (e.g., type of wavelets,
number of filters, their frequency overlaps, etc.). 

Finally, we note that our optimization strategy Eq.~\eqref{eqn:opt} should work in principle with any other learning model as long as it outputs the class probability $p_k(\bx)$ for a given input signal $\bx \in \Rf^\ndim$, e.g.,
a DL model equipped with ``softmax'' output units~\cite[Sec.\ 6.2.2.3]{Goodfellow-et-al-2016}.

\section*{Acknowledgment}
This research was partially supported by the US National Science Foundation grants DMS-1912747 and CCF-1934568 as well as the US Office of Naval Research grant N00014-20-1-2381.

\end{document}